\pdfoutput=1

\documentclass[11pt]{article}
\usepackage[dvipsnames]{xcolor}

\usepackage{acl}

\usepackage{times}
\usepackage{latexsym}

\usepackage[T1]{fontenc}

\usepackage[utf8]{inputenc}

\usepackage{microtype}

\usepackage{booktabs}

\usepackage{natbib}
\usepackage{graphicx}
\usepackage{amsmath, amsfonts, amssymb}
\usepackage{geometry}
\usepackage{booktabs}
\usepackage{xurl}
\usepackage{enumitem}
\usepackage{dblfloatfix}
\usepackage{comment}
\usepackage{placeins}
\usepackage{enumitem}

\makeatletter
\renewcommand{\paragraph}{%
  \@startsection{paragraph}{4}%
  {\z@}{0ex \@plus 0ex \@minus 0ex}{-1em}%
  {\normalfont\normalsize\bfseries}%
}
\makeatother

\title{The Irrationality of Neural Rationale Models}

\author{Yiming Zheng \qquad Serena Booth \qquad Julie Shah \qquad Yilun Zhou \\
  MIT CSAIL\\
  \texttt{yimingz@mit.edu \{serenabooth,julie\_a\_shah,yilun\}@csail.mit.edu}}

\begin{document}
\maketitle
\begin{abstract}
Neural rationale models are popular for interpretable predictions of NLP tasks. In these, a selector extracts segments of the input text, called \textit{rationales}, and passes these segments to a classifier for prediction. Since the rationale is the only information accessible to the classifier, it is plausibly \textit{defined} as the explanation. Is such a characterization unconditionally correct? In this paper, we argue to the contrary, with both philosophical perspectives and empirical evidence suggesting that rationale models are, perhaps, less rational and interpretable than expected. We call for more rigorous evaluations of these models to ensure desired properties of interpretability are indeed achieved. The code for our experiments is at \url{https://github.com/yimingz89/Neural-Rationale-Analysis}. 
\end{abstract}

\section{Introduction}

As machine learning models are increasingly used in high-stakes domains, understanding the reasons for a prediction becomes more important, especially when the model is a black-box such as a neural network. While many \textit{post-hoc} interpretability methods have been developed for models operating on tabular, image, and text data~\citep{simonyan2013deep, ribeiro2016should, feng2018pathologies}, their faithfulness are often questioned \citep{adebayo2018sanity, rudin2019stop, zhou2021feature}. 

\begin{figure}[t]
  \centering
   \includegraphics[width=0.9\linewidth]{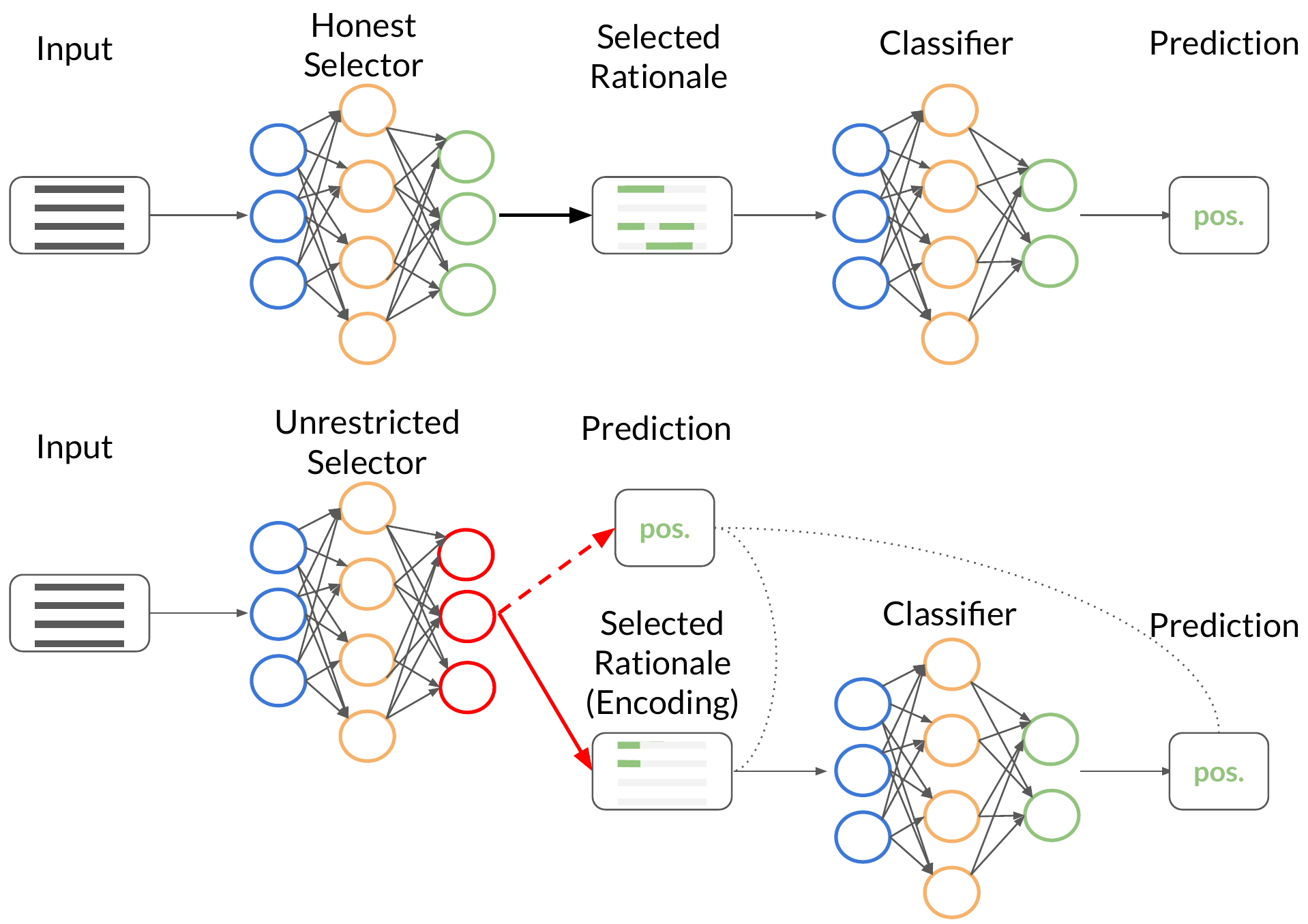}
   \caption{Top: an honest neural rationale model. We seek to understand the selector's process (the bold arrow), which should select words and phrases as rationale in an unbiased way, leaving the prediction to the classifier which receives this rationale. Bottom: a failure case of neural rationale models. As discussed in Section \ref{sec:perspective}, an unrestricted selector may be able to make its own (relatively accurate) prediction, and ``pass'' it to the classifier via encoding it in the selected rationale.}
   \label{fig:main}
\end{figure}

With no resolution in sight for explaining black box models, \textit{inherently interpretable} models, which self-explain while making decisions, are often favored. Neural rationale models, shown in Figure \ref{fig:main} (top), are the most popular in NLP~\citep{lei-etal-2016-rationalizing, bastings-etal-2019-interpretable, yu2019rethinking, jain2020learning}: in them, a selector processes the input text, extracts segments (i.e. \textit{rationale}) from it, and sends \textit{only} the rationale to the predictor. Since the rationale is the only information accessible to the predictor, it arguably serves as the \textit{explanation} for the prediction.

While the bottleneck structure defines a causal relationship between rationale and prediction, we caution against equating this structure with inherent interpretability without additional constraints. Notably, if both the selector and the classifier are sufficiently flexible function approximators (e.g. neural networks), the bottleneck structure provides \textit{no} intrinsic interpretability as the selector and classifier may exploit imperceptible messages, as shown in Figure \ref{fig:main} (bottom).

We perform a suite of empirical analyses to demonstrate how rationales lack interpretability. Specifically, we present modes of instability of the rationale selection process under minimal and meaning-preserving sentence perturbations on the Stanford Sentiment Treebank \citep[SST, ][]{socher-etal-2013-recursive} dataset. Through a user study, we further show that this instability is poorly understood by people---even those with advanced machine learning knowledge. 
We find that the exact form of interpretability induced by neural rationale models, if any, is not clear. As a community, we must critically reflect on the interpretability of these models, and perform rigorous evaluations about any and all claims of interpretability going forward.

\section{Related Work}
Most interpretability efforts focus on \textit{post-hoc} interpretation. For a specific input, these methods generate an explanation by analyzing model behaviors such as gradient \citep{simonyan2013deep, sundararajan2017axiomatic} or prediction on perturbed \citep{ribeiro2016should, lundberg2017unified} or reduced \citep{feng2018pathologies} inputs. However, evaluations of these methods highlight various problems. For example, \citet{adebayo2018sanity} showed that many methods can generate seemingly reasonable explanations even for random neural networks. \citet{zhou2021feature} found that many methods fail to identify features known to be used by the model. \citet{zhou22exsum} share the same principles as us, but also focus on general post-hoc interpretations of arbitrary black-box models, while we focus on neural rationale models.

By contrast, neural rationale models are largely deemed \textit{inherently interpretable} and thus do not require \textit{post-hoc} analysis. At a high level, a model has a selector and a classifier. For an input sentence, the selector first calculates the \textit{rationale} as excerpts of the input, and then the classifier makes a prediction from \textit{only} the rationale. Thus, the rationale is often \textit{defined} as the explanation due to this bottleneck structure. The non-differentiable rationale selection prompts people to train the selector using policy gradient \citep{lei-etal-2016-rationalizing, yu2019rethinking} or continuous relaxation \citep{bastings-etal-2019-interpretable}, or directly use a pre-trained one \citep{jain2020learning}. 

While rationale models have mostly been subject to less scrutiny, some evaluations have been carried out. \citet{yu2019rethinking} proposed the notions of comprehensiveness and sufficiency for rationales, advocated as standard evaluations in the ERASER \citep{deyoung2019eraser} dataset. \citet{zhou2021feature} noted that training difficulty, especially due to policy gradient, leads to selection of words known to not influence the label in the data generative model. Complementing these evaluations and criticisms, we argue from additional angles to be wary of interpretability claims for rationale models, and present experiments showing issues with existing models. 

Most related to our work, \citet{DBLP:journals/corr/abs-2006-01067} mention a Trojan explanation and dominant selector as two failure modes of rationale models. We pinpoint the same root cause of a non-understandable selector in Section \ref{sec:perspective}. However, they favor rationales generated \textit{after} the prediction, while we will argue for rationales being generated \textit{prior} to the prediction. Also, in their discussion of contrastive explanations, their proposed procedure runs the model on out-of-distribution data (sentence with some tokens masked), potentially leading to arbitrary predictions due to extrapolation, a criticism also argued by \citet{DBLP:journals/corr/abs-1806-10758}.
\section{Philosophical Perspectives}
\label{sec:perspective}
In neural rationale models, the classifier prediction causally results from the selector rationale, but does this property automatically equate rationale with explanation? We first present a ``failure case.'' For a binary sentiment classification, we first train a (non-interpretable) classifier $c'$ that predicts on the whole input. Then we \textit{define} a selector $s'$ that selects the first word of the input if the prediction is positive, or the first two words if the prediction is negative. Finally, we train a classifier $c$ to imitate the prediction of $c'$ but from the rationale. The $c'\rightarrow s'\rightarrow c$ model should achieve best achievable accuracy, since the actual prediction is made by the unrestricted classifier $c'$ with full input access. Can we consider the rationale as explanation? No, because the rationale selection depends on, and is as (non-)interpretable as, the black-box $c'$. This failure case is shown in Figure \ref{fig:main} (bottom). Recently proposed introspective training \citep{yu2019rethinking} could not solve this problem either, as the selector can simply output the comprehensive rationale along with the original cue of first one or two words, with only the latter used by the classifier\footnote{In fact, the extended rationale \textit{helps} disguise the problem by appearing as much more reasonable. }. In general, a sufficiently powerful selector can make the prediction at selection time, and then pass this prediction via some encoding in the selected rationale for the classifier to use.

To hide the ``bug,'' consider now $s'$ selecting the three most positive or negative words in the sentence according to the $c'$ prediction (as measured by embedding distance to a list of pre-defined positive/negative words). This model would seem very reasonable to a human, yet it is non-interpretable for the same reason. To recover a ``native'' neural model, we could train a selector $s$ to imitate $c'\rightarrow s'$ via teacher-student distillation \citep{hinton2015distilling}, and the innocent-looking $s\rightarrow c$ rationale model remains equally non-interpretable. 

Even without the explicit multi-stage supervision above, a sufficiently flexible selector $s$ (e.g. a neural network) can implicitly learn the $c'\rightarrow s'$ model and essentially control the learning of the classifier $c$, in which case the bottleneck of succinct rationale affords no benefits of interpretability. So why does interpretability get lost (or fail to emerge)? The issue arises from not understanding the \textit{rationale selection process}, i.e. selector $s$. If it is well-understood, we could determine its true logic to be $c'\rightarrow s'$ and reject it. Conversely, if we cannot understand \textit{why} a particular rationale is selected, then accepting it (and the resulting prediction) at face value is not really any different from accepting an end-to-end prediction at face value. 

In addition, the selector-classifier decomposition suggests that the selector should be an ``unbiased evidence collector'', i.e. scanning through the input and highlighting all relevant information, while the classifier should deliberate on the evidence for each class and make the decision. Verifying this role of the selector would again require its interpretability. 

Finally, considering the rationale model as a whole, we could also argue that the rationale selector \textit{should} be interpretable. It is already accepted that the classifier can remain a black-box. If the selector is also not interpretable, then exactly what about the model is interpretable? 

\begin{figure}[t]
    \centering
   \includegraphics[width=0.8\linewidth]{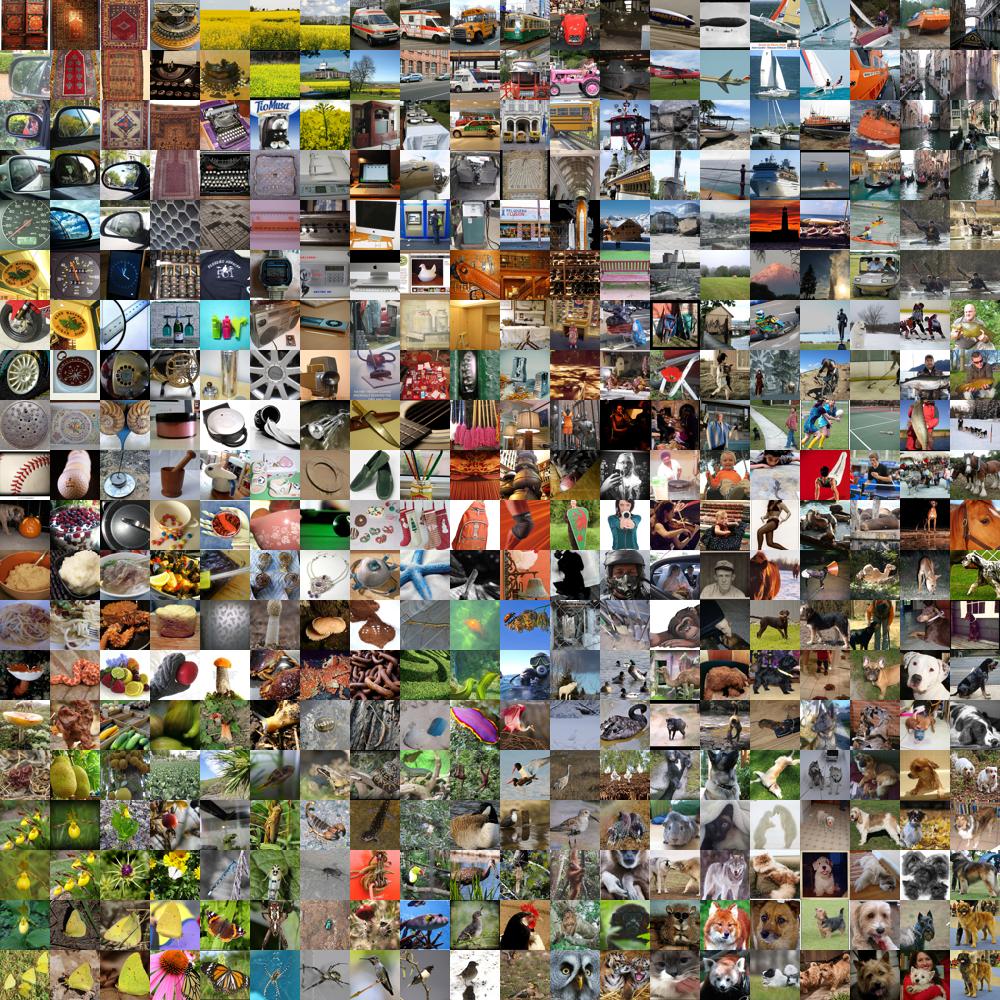}
    \caption{Embedding space visualization of an ImageNet classifier. Image from \url{https://cs.stanford.edu/people/karpathy/cnnembed/}. }
    \label{fig:imagenet-embedding}
\end{figure}

Architecturally, we can draw an analogy between the rationale in rationale models and the embedding representation in a typical end-to-end classifier produced at the penultimate layer. A rationale is a condensed feature extracted by the selector and used by the classifier, while, for example in image models, the image embedding is the semantic feature produced by the feature extractor and used by the final layer of linear classifier. Furthermore, both of them exhibit some interpretable properties: rationales represent the ``essence'' of the input, while the image embedding space also seems semantically organized (e.g. Figure \ref{fig:imagenet-embedding} showing ImageNet images organized in the embedding space). However, this embedding space is rarely considered on its own as the explanation for a prediction, exactly because the feature extractor is a black-box. Similarly, the rationales by default should not qualify as the explanation either, despite its textual nature.

Finally, from a practical perspective, explanations should help humans understand the model's input-output behavior. Such a purpose is fulfilled when the human understands not only why an explanation leads to an output, but also \textit{how} the explanation is generated from the input in the first place. Our emphasis on understanding the rationale selection process fulfills the latter requirement. Such a perspective is also echoed by \citet{pruthi2020}, who argued that the practical utility of explanations depends crucially on human's capability of understanding how they are generated.

\section{Empirical Investigation}
\label{sec:experiment}
As discussed above, truly interpretable rationale models require an understanding of the rationale selection process. However, since the selector is a sequence-to-sequence model, for which there is no standard methods for interpretability, we focus on a ``necessary condition'' setup of understanding the input-output behavior of the model in our empirical investigation. Specifically, we investigate rationale selection changes in response to meaning-preserving non-adversarial perturbation of individual words in the input sentence.

\subsection{Setup}
On the 5-way SST dataset \citep{socher-etal-2013-recursive}, we trained two rationale models, a continuous relaxation (CR) model \citep{bastings-etal-2019-interpretable} and a policy gradient (PG) model \citep{lei-etal-2016-rationalizing}. The PG model directly generates binary (i.e. hard) rationale selection. The CR model uses a $[0, 1]$ continuous value to represent selection and scales the word embedding by this value. Thus, we consider a word being selected as rationale if this value is non-zero.
Our CR model achieves 47.3\% test accuracy with 24.9\% rationale selection rate (i.e. percentage of words in the input selected as rationale), and PG model 43.3\% test accuracy with 23.1\% rationale selection rate, consistent with those obtained by \citet[][Figure 4]{bastings-etal-2019-interpretable}. Additional details are in Appendix~\ref{app:detail-setup}.

\subsection{Sentence Perturbation Procedure}
\label{section:perturbation}
The perturbation procedure changes a noun, verb, or adjective as parsed by NLTK\footnote{i.e. NN, NNS, VB, VBG, VBD, VBN, VBP, VBZ, and JJ} \citep{journals/corr/cs-CL-0205028} with two requirements. First, the new sentence should be natural (e.g., ``I \textit{observed} a movie'' is not). Second, its meaning should not change (e.g. adjectives should not be replaced by antonyms). 

For the first requirement, we \texttt{[MASK]} the candidate word and use the pre-trained BERT \citep{devlin-etal-2019-bert} to propose 30 new choices. For the second requirement, we compute the union of words in the WordNet synset associated with each definition of the candidate words \citep{WordNetBook}. If the two sets share no common words, we mark the candidate invalid. Otherwise, we choose the top BERT-predicted word as the replacement.

We run this procedure on the SST test set, and construct the perturbed dataset from all valid replacements of each sentence. Table \ref{table:example} lists some example perturbations (more in Appendix \ref{app:perturbation-examples}). Table \ref{table:preds} shows the label prediction distribution on the original test set along with changes due to perturbation in parentheses, and confirms that the change is overall very small. Finally, a human evaluation checks the perturbation quality, detailed in Appendix \ref{app:pert-user-study}.
For 100 perturbations, 91 were rated to have the same sentiment value. Furthermore, on all 91 sentences, the same rationale is considered adequate to support the prediction after perturbation as well.

\begin{table}[h!]
\resizebox{\columnwidth}{!}{
\begin{tabular}{p{3.6in}}\toprule 
    A pleasurably jacked-up \textbf{piece}/\textit{slice} of action moviemaking .\\\midrule
    The \textbf{use}/\textit{usage} of CGI and digital ink-and-paint make the thing look really slick . \\\bottomrule
\end{tabular}
} 
\caption{Sentence perturbation examples, with the original word in \textbf{bold} replaced by the word in \textit{italics}. }
\vspace{-0.01in}
\label{table:example}
\end{table}

\begin{table}[h!]
\centering
\resizebox{\columnwidth}{!}{
\begin{tabular}{c|c|c|c|c|c}\toprule
     & 0 & 1 & 2 & 3 & 4 \\\midrule
    CR & 8.0 (-0.6) & 41.5 (+1.5) & 8.9 (-0.4) & 28.6 (+1.0) & 13.0 (-1.5) \\
    PG & 8.6 (-1.6) & 40.7 (-1.3) & 1.6 (-0.2) & 33.9 (+5.0) & 15.2 (-1.9) \\\bottomrule
\end{tabular}
} 
\caption{The percentage of predicted labels on the original test set, as well as the differences to the that on the perturbation sentences in parentheses.}
\vspace{-0.05in}
\label{table:preds}
\end{table}

\subsection{Results}
Now we study the effects of perturbation on rationale selection change (i.e. an originally selected word getting unselected or vice versa). We use only perturbations that maintain the model prediction, as in this case, the model is expected to use the same rationales according to human evaluation. 

\paragraph{Qualitative Examples} Table \ref{table:rationale} shows examples of rationale changes under perturbation (more in Appendix \ref{app:rationale-change-examples}). Indeed, minor changes can induce nontrivial rationale change, sometimes far away from the perturbation location. Moreover, there is no clear relationship between the words with selection change and the perturbed word. 

\begin{table}[h!]
\resizebox{\columnwidth}{!}{
\begin{tabular}{c|p{3.6in}}\toprule
    PG & The \textbf{story}/\textcolor{blue}{\textit{narrative}} loses its \textcolor{blue}{bite} in a \textcolor{Green}{last-minute} happy \textcolor{Green}{ending} that 's even \textcolor{Green}{less plausible} than the rest of the picture .\\\midrule
    PG & A \textcolor{Green}{pleasant ramble} through the \textcolor{Green}{sort} of \textcolor{red}{idoosyncratic} \textcolor{Green}{terrain} that Errol Morris \textbf{has}/\textit{have} \textcolor{red}{often dealt} with ... it does possess a \textcolor{Green}{loose} , \textcolor{Green}{lackadaisical charm} .\\\midrule 
    CR & \textcolor{blue}{I love} the way that it took \textcolor{Green}{chances} and really \textcolor{Green}{asks} you to take these \textbf{great}/\textit{big} \textcolor{red}{leaps} of \textcolor{Green}{faith} and \textcolor{Green}{pays off} .\\\midrule
    CR & Legendary \textcolor{blue}{Irish} \textcolor{red}{\textbf{writer}}/\textit{author} \textcolor{Green}{Brendan} Behan 's \textcolor{blue}{memoir} , \textcolor{Green}{Borstal} Boy , has been \textcolor{Green}{given} a \textcolor{Green}{loving} \textcolor{blue}{screen} \textcolor{Green}{transferral} . \\\bottomrule
\end{tabular}
}
\caption{Rationale change example. Words selected in the original only, perturbed only, and both are shown in \textcolor{red}{red}, \textcolor{blue}{blue}, and \textcolor{Green}{green}, respectively.}
\vspace{-0.1in}
\label{table:rationale}
\end{table}

\paragraph{Rationale Change Freq.}
Quantitatively, we first study how often rationales change. Table \ref{table:selection-changes} shows the count frequency of selection changes. Around 30\% (non-adversarial) perturbations result in rationale change (i.e. non-zero number of changes). Despite better accuracy, the CR model is less stable and calls for more investigation into its selector. 
\begin{table}[h!]
\centering
\resizebox{\columnwidth}{!}{
\begin{tabular}{c|c|c|c|c|c|c}\toprule
    \# Change & 0 & 1 & 2 & 3 & 4 & $\geq 5$ \\\midrule
    CR & 66.5\% & 25.5\% & 6.8\% & 1.0\% & 0.1\% & 0.1\% \\
    PG & 77.4\% & 21.4\% & 1.1\% & 0.1\% & 0\% & 0\% \\\bottomrule
\end{tabular}
}
\caption{Frequency of number of selection changes. }
\vspace{-0.025in}
\label{table:selection-changes}
\end{table}


\begin{figure}[t]
  \centering
   \includegraphics[width=0.9\linewidth]{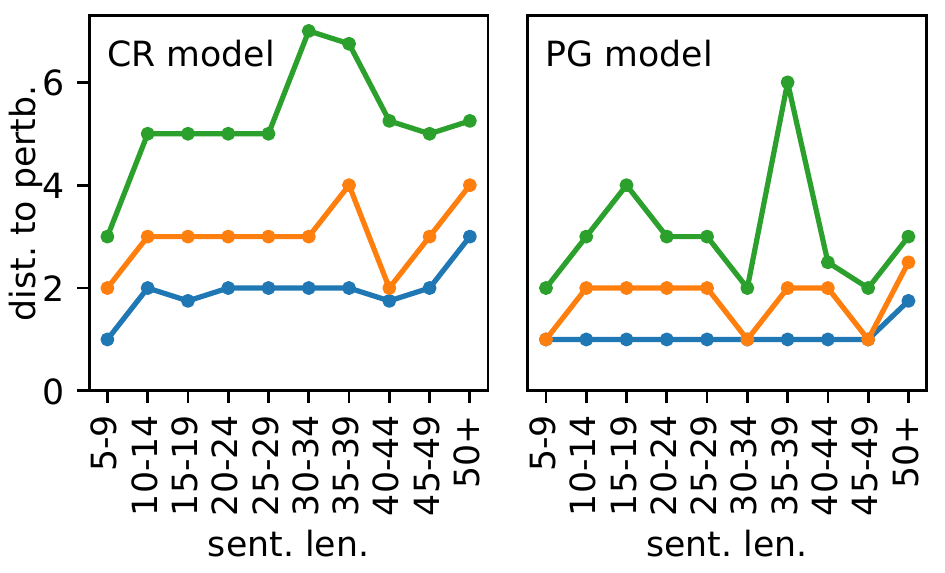}
   \caption{Scatter plots showing three quartiles of distance between indirect rationale change to perturbation, grouped by sentence length.}
   \label{fig:dist}
\end{figure}

\begin{figure}[t]
  \centering
   \includegraphics[width=0.9\linewidth]{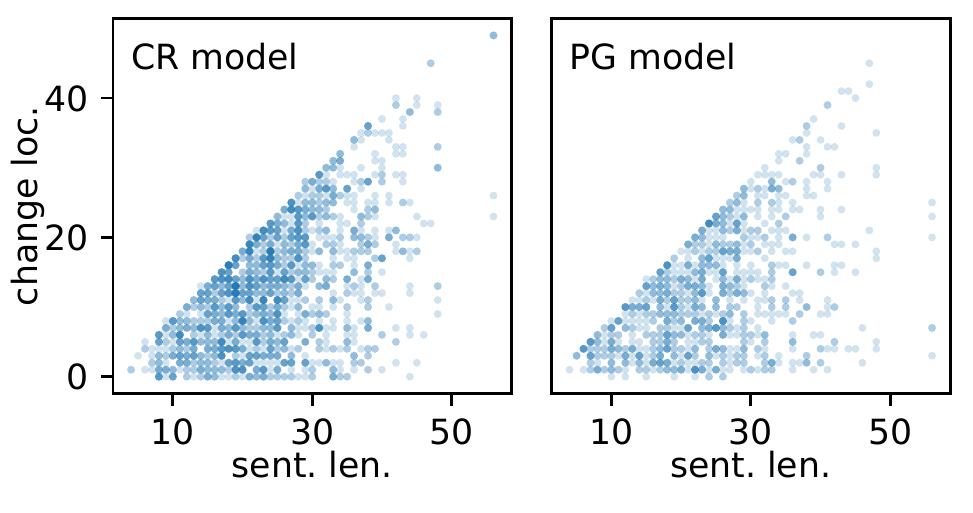}
   \caption{Locations of all selection changes, with each one shown as a dot. }
   \label{fig:index}
\end{figure}

\begin{figure}[t]
  \centering
   \includegraphics[width=0.65\linewidth]{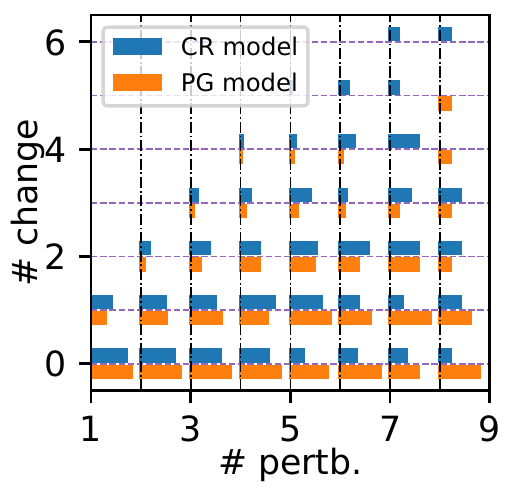}
   \caption{For sentences with a certain number of valid perturbations, the corresponding column of bar chart shows the count frequency of perturbations that result in any rationale change.}
   \label{fig:stability}
\end{figure}

\paragraph{Locations of Selection Change}
Where do these changes occur? 29.6\% and 78.3\% of them happen at the perturbed word for the CR and PG models respectively. For the CR model, over 70\% of rationale changes are due to replacements of \textit{other} words; this statistic is especially alarming. For these indirect changes, Figure \ref{fig:dist} shows the quartiles of distances to the perturbation for varying sentence lengths. They are relatively constant throughout, suggesting that the selection uses mostly local information. However, the ``locality size'' for CR is about twice as large, and changes often occur five or more words away from the perturbation.

We also compute the (absolute) location of the rationale changes, as plotted in Figure \ref{fig:index}, where each dot represents an instance. The rationale changes are distributed pretty evenly in the sentence, making it hard to associate particular perturbation properties to the resulting selection change location. 

\paragraph{Sentence-Level Stability}
Are all the rationale changes concentrated on a few sentences for which every perturbation is likely to result in a change, or are they spread out across many sentences? We measure the \textit{stability} of a sentence by the number of perturbations inducing rationale changes.
Obviously, a sentence with more valid perturbations is likely to also have more change-inducing ones, so we plot the frequency of sentences with a certain stability value separately for different total numbers of perturbations in Figure \ref{fig:stability}. There are very few highly unstable sentences, suggesting that the selection change is a common phenomenon to most of the sentences, further adding to the difficulty of a comprehensive understanding of the selector. 

\begin{table*}[h!]
\centering
\resizebox{\textwidth}{!}{
\begin{tabular}{c|c|c|c|c|c|c|c}\toprule
    POS (frequency) & noun (19.2\%) & verb (14.3\%) & adj. (10.1\%) & adv. (5.8\%) & proper n. (4.4\%) & pron. (4.9\%) & other (41.3\%) \\\midrule
    CR change / all & 37.1\% / 34.3\% & 21.9\% / 16.0\% & 14.2\% / 24.8\% & 8.9\% / 11.3\% & 3.5\% / 5.8\% & 2.5\% / 1.0\% & 11.9\% / 6.8\%\\
    PG change / all & 42.7\% / 33.6\% & 30.2\% / 16.6\% & 20.6\% / 30.6\% & 2.4\% / 12.9\% & 1.8\% / 3.4\% & 0.4\% / 0.5\% & 1.9\% / 2.4\% \\\bottomrule
\end{tabular}
}
\caption{Part of speech (POS) statistics. The top row shows the POS composition of the test set sentences. The bottom two rows show POS composition for changed rationale words and for all rationale words.}
\vspace{-0.1in}
\label{table:pos-changes}
\end{table*}

\paragraph{Part of Speech Analysis}
Our final automated analysis studies the part-of-speech (POS) composition of selection changes. As Table \ref{table:pos-changes}  shows, adjectives and adverbs are relatively stable, as expected because they encode most sentiments. By contrast, nouns and verbs are less stable, probably because they typically represent factual ``content'' that is less important for prediction. The CR model is especially unstable for other POS types such as determiner and preposition. Overall, the instability adds to the selector complexity and could even function as subtle ``cues'' described in Section \ref{sec:perspective}. 

\paragraph{User Study on Selector Understanding} While the automated analyses reveal potential obstacles to selector understanding, ultimately the problem is the lack of understanding by users. The most popular way to understand a model is via input-output examples \citep{ribeiro2020beyond, booth2021bayes}, and we conduct a user study in which we ask participants (grad students with ML knowledge) to match rationale patterns with sentences before and after perturbation on 20 instances, after observing 10 true model decisions (details in Appendix \ref{app:rationale-user-study}). Unsurprisingly, participants get 45 correct out of 80 pairs, basically at the random guess level, even as some participants use reasons related to grammar and atypical word usage (which are apparently ineffective), along with ``lots of guessing''. This result confirms the lack of selector understanding even under minimal perturbation, indicating more severity for completely novel inputs. 

\section{Conclusion}
We argue against the commonly held belief that rationale models are inherently interpretable by design. We present several reasons, including a counter-example showing that a reasonable-looking model could be as non-interpretable as a black-box. These reasons imply that the missing piece is an understanding of the \textit{rationale selection process} (i.e. the selector). We also conduct a (non-adversarial) perturbation-based study to investigate the selector of two rationale models, in which automated analyses and a user study confirm that they are indeed hard to understand. In particular, the higher-accuracy model (CR) fares worse in most aspects, possibly hinting at the performance-interpretability trade-off \citep{gunning2019darpa}. These results point to a need for more rigorous analysis of interpretability in neural rationale models.

\bibliography{anthology,custom}
\bibliographystyle{acl_natbib}

\newpage

\onecolumn
\appendix

\section{Additional Details on the Experimental Setup}
\label{app:detail-setup}

\subsection{Training}
The models we train are as implemented in \citep{bastings-etal-2019-interpretable}. The hyperparameters we use are 30 percent for the word selection frequency when training the CR model and $L_0$ penalty weight 0.01505 when training the PG model. Training was done on a MacBook Pro with a 1.4 GHz Quad-Core Intel Core i5 processor and 8 GB 2133 MHz LPDDR3 memory. The training time for each model was around 15 minutes. There are a total of 7305007 parameters in the CR model and 7304706 parameters in the PG model. The hyperparameter for the CR model is the word selection frequency, ranging from 0\% to 100\%, whereas the hyperparameter for the PG model is the $L_0$ penalty weight which is a nonnegative real number (for penalizing gaps in selections).

These hyperparameter were configured with the goal that both models would select a similar fraction of total words as rationale. This was done manually. Only one CR model was trained (with the word selection frequency set to 30 percent). Then, a total of 7 PG models were trained, with $L_0$ penalty weight ranging from 0.01 to 0.025. Then, the closest matching result to the CR model in terms of word selection fraction, which was an $L_0$ penalty of 0.01505, was used. 

The CR model (with 30\% word selection frequency) achieves a 47.3\% test accuracy with a 24.9\% rationale selection rate, and the PG model (with $L_0$ penalty of 0.01505) achieves a 43.3\% test accuracy with a 23.1\% selection rate, consistent with those obtained by \citet[][Figure 4]{bastings-etal-2019-interpretable}. The CR model achieves a validation accuracy of 46.0\% with a 25.1\% rationale selection rate, and the PG model achieves a 41.1\% validation accuracy with a 22.9\% selection rate, comparable to the test results. 

\subsection{Dataset}
We use the Stanford Sentiment Treebank \citep[SST, ][]{socher-etal-2013-recursive} dataset with the exact same preprocessing and train/validation/test split as given by \citet{bastings-etal-2019-interpretable}. There are 11855 total entries (each are single sentence movie reviews in English), split into a training size of 8544, a validation size of 1101, and a test size of 2210. The label distribution is 1510 sentences of label 0 (strongly negative), 3140 of label 1 (negative), 2242 of label 2 (neutral), 3111 of label 3 (positive), and 1852 of label 4 (strongly positive). We use this dataset as is, and no further pre-processing is done. The dataset can be downloaded from the code provided by \citet{bastings-etal-2019-interpretable}.

\subsection{Sentence Perturbation}
The data perturbation was done on the same machine with specs described in Appendix \ref{app:detail-setup}. This procedure was done once and took around an hour. This perturbation was an automated procedure using the BERT and WordNet synset intersection as a heuristic for word substitutions. As a result, we did not collect any new data which requires human annotation or other work.

\newpage
\section{Additional Examples of Sentence Perturbation}
\label{app:perturbation-examples}

Table \ref{table:appendix-examples} shows ten randomly sampled perturbations.

\begin{table}[h!]
\resizebox{\columnwidth}{!}{
\begin{tabular}{p{8in}}\toprule
    There are weird resonances between actor and \textbf{role}/\textit{character} here , and they 're not exactly flattering .\\\midrule 
    A loving \textbf{little}/\textit{short} film of considerable appeal .\\\midrule
    The film is really not so \textbf{much}/\textit{often} bad as bland .\\\midrule
    A cockamamie tone poem pitched precipitously between swoony lyricism and violent catastrophe ... the most aggressively nerve-wracking and screamingly neurotic romantic comedy in \textbf{cinema}/\textit{film} history .\\\midrule
    Steve Irwin 's method is Ernest Hemmingway at accelerated speed and \textbf{volume}/\textit{mass} .\\\midrule
    The movie addresses a hungry need for PG-rated , nonthreatening family \textbf{movies}/\textit{film} , but it does n't go too much further .\\\midrule
    ... the last time I saw a theater full of people constantly checking their \textbf{watches}/\textit{watch} was during my SATs .\\\midrule
    Obvious politics and rudimentary animation reduce the \textbf{chances}/\textit{chance} that the appeal of Hey Arnold !\\\midrule
    Andy Garcia enjoys one of his richest roles in years and Mick Jagger gives his best \textbf{movie}/\textit{film} performance since , well , Performance .\\\midrule
    Beyond a handful of mildly amusing lines ... there just \textbf{is}/\textit{be} n't much to laugh at .
    \\\bottomrule
\end{tabular}
} 
\caption{Ten randomly sampled sentence perturbation examples given in a user study, with the original word shown in \textbf{bold} replaced by the word in \textit{italics}. }
\label{table:appendix-examples}
\end{table}

\section{Description of the Human Evaluation of Data Perturbation}
\label{app:pert-user-study}
We recruited five graduate students with ML experience (but no particular experience with interpretable ML or NLP), and each participant was asked to answer questions for 20 sentence perturbations, for a total of 100 perturbations. An example question is shown below:

\begin{center}
\fbox{
\begin{minipage}{0.8\textwidth}
The original sentence (a) and the perturbed sentence (b), as well as the selected rationale on the original sentence (in bold) are:
\vspace{-0.05in}
\begin{enumerate}[label=\alph*, parsep=-0.05in]
    \item There \textbf{are weird resonances} between actor and \underline{role} here , and they \textbf{'re} not \textbf{exactly flattering} .
    \item There are weird resonances between actor and \underline{character} here , and they 're not exactly flattering .
\end{enumerate}
\vspace{-0.1in}
The original prediction is: negative. 
\vspace{-0.05in}
\begin{enumerate}[parsep=-0.05in]
    \item Should the prediction change, and if so, in which way:
    \item If yes:
    \begin{enumerate}
        \item Does the changed word need to be included or removed from the rationale?
        \item Please highlight the new rationale in red directly on the new sentence. 
    \end{enumerate}
\end{enumerate}
\end{minipage}
}
\end{center}
The study takes less than 15 minutes, is conducted during normal working hours with participants being grad students on regular stipends, and is uncompensated. 

\newpage

\section{Additional Rationale Change Examples}
\label{app:rationale-change-examples}
Table \ref{table:appendix-rationale-change-examples} shows additional rationale change examples.

\begin{table}[h!]
\resizebox{\columnwidth}{!}{
\begin{tabular}{c|p{6.5in}}\toprule
    PG & This \textcolor{Green}{delicately} observed \textbf{story}/\textcolor{blue}{\textit{tale}} , \textcolor{Green}{deeply} \textcolor{blue}{felt} and \textcolor{Green}{masterfully stylized} , is a \textcolor{Green}{triumph} for its \textcolor{Green}{maverick} director. \\\midrule 
    PG & Biggie and Tupac is \textcolor{Green}{so single-mindedly daring} , it \textcolor{red}{\textbf{puts}}/\textit{put} \textcolor{red}{far} \textcolor{Green}{more polished} \textcolor{red}{documentaries} to \textcolor{Green}{shame}.\\\midrule
    PG & \textcolor{Green}{Somewhere} short of Tremors on the modern B-scene : neither as \textcolor{Green}{funny} nor as \textcolor{Green}{clever} , though an \textcolor{Green}{agreeably unpretentious} way to \textcolor{red}{\textbf{spend}}/\textit{pass} \textcolor{blue}{ninety} minutes .\\\midrule
    PG & The film overcomes the regular \textcolor{Green}{minefield} of \textcolor{Green}{coming-of-age cliches} with \textcolor{red}{\textbf{potent}}/\textit{strong} \textcolor{red}{doses} of \textcolor{Green}{honesty} and \textcolor{Green}{sensitivity} .\\\midrule
    PG & As expected , Sayles ' \textcolor{Green}{smart wordplay} and \textcolor{Green}{clever plot contrivances are} as \textcolor{Green}{sharp} as ever , though they may be \textcolor{Green}{overshadowed} by \textcolor{blue}{some} \textbf{strong}/\textcolor{blue}{\textit{solid}} \textcolor{Green}{performances} .\\\midrule
    CR & The \textcolor{blue}{animated} \textcolor{Green}{subplot keenly} \textcolor{blue}{depicts} the inner \textcolor{red}{\textbf{struggles}}/\textit{conflict} of our adolescent heroes - \textcolor{Green}{insecure} , \textcolor{Green}{uncontrolled} , \textcolor{blue}{and} \textcolor{Green}{intense} .\\\midrule 
    CR & \textcolor{Green}{Funny} and , at \textcolor{Green}{times} , \textcolor{Green}{poignant} , the \textcolor{blue}{film} from director George \textcolor{Green}{Hickenlooper} \textcolor{red}{all} \textbf{takes}/\textit{take} place in \textcolor{Green}{Pasadena} , \textcolor{red}{``} a \textcolor{Green}{city} where \textcolor{blue}{people} \textcolor{Green}{still} read . ''\\\midrule 
    CR & It would be hard to think of a recent movie that \textbf{has}/\textcolor{blue}{\textit{have}} \textcolor{red}{worked} this \textcolor{Green}{hard} to \textcolor{red}{achieve} this \textcolor{red}{little} fun. \\\midrule 
    CR & This \textcolor{Green}{road} movie \textbf{gives}/\textit{give} \textcolor{red}{you} emotional \textcolor{red}{whiplash} , and you 'll be \textcolor{Green}{glad} \textcolor{blue}{you} went along for the \textcolor{Green}{ride} .\\\midrule
    CR & If nothing else , this movie introduces a \textcolor{Green}{promising , unusual} \textcolor{red}{\textbf{kind}}/\textit{form} of \textcolor{blue}{psychological horror} .
    \\\bottomrule
\end{tabular}
} 
\caption{Additional rationale change example. Words selected in the original only, perturbed only, and both are shown in \textcolor{red}{red}, \textcolor{blue}{blue}, and \textcolor{Green}{green}, respectively.}
\label{table:appendix-rationale-change-examples}
\end{table}

\section{Description of the User Study on Rationale Change}
\label{app:rationale-user-study}
Participants were first given 10 examples of rationale selections (shown in bold) on the original and perturbed sentence pair made by the model, with one shown below: 

\begin{center}
\fbox{
\begin{minipage}{0.8\textwidth}
orig: \textbf{Escapism} in its \textbf{purest} \underline{form} .

pert: \textbf{Escapism} in its \textbf{purest \underline{kind}} .
\end{minipage}
}
\end{center}
Then, they were presented with 20 test questions, where each question had two rationale assignments, one correct and one mismatched, and they were asked to determine which was the correct rationale assignment. An example is shown below:

\begin{center}
\fbox{
\begin{minipage}{0.8\textwidth}
\vspace{0.1in}
\begin{enumerate}[label=\alph*, parsep=-0.05in, leftmargin=0.15in]
    \item orig: \textbf{Benefits} from a \textbf{\underline{strong} performance} from Zhao , but it 's Dong Jie 's \textbf{face} you \textbf{remember} at the end .\\
    pert: Benefits from a \textbf{\underline{solid}} performance from Zhao , but it 's Dong Jie 's \textbf{face} you \textbf{remember} at the end
    \item orig: Benefits from a \textbf{\underline{strong}} performance from Zhao , but it 's Dong Jie 's \textbf{face} you \textbf{remember} at the end .\\
    pert: \textbf{Benefits} from a \textbf{\underline{solid} performance} from Zhao , but it 's Dong Jie 's \textbf{face} you \textbf{remember} at the end
\end{enumerate}
\vspace{-0.1in}
In your opinion, which pair (a or b) shows the actual rationale selection by the model? 
\end{minipage}
\vspace{0.05in}
}
\end{center}
In the end, we ask the participants the following question for any additional feedback. 
\begin{center}
    \fbox{
    \begin{minipage}{0.8\textwidth}
    Please briefly describe how you made the decisions (which could include guessing), and your impression of the model’s behavior. 
    \end{minipage}
    }
\end{center}
The study takes less than 15 minutes, is conducted during normal working hours with participants being grad students on regular stipends, and is uncompensated. 

\end{document}